%% file: main.tex
\documentclass[journal]{IEEEtran}

\usepackage{graphicx} 
\usepackage{tabularx}
\usepackage{array}

\usepackage{amsmath,amsfonts,bm,bbm,enumitem}

\allowdisplaybreaks
\usepackage[mathscr]{euscript}
\usepackage{color}
\usepackage{amsthm}
\usepackage{graphicx}
\usepackage{url}
\usepackage{multirow,hhline}
\usepackage{dblfloatfix}
\usepackage{amssymb}
\usepackage{mathalfa} 
\usepackage{caption}

\usepackage{algorithm} 
\usepackage[noend]{algpseudocode} 
\usepackage{amsmath} 
\usepackage{amssymb} 
 
\newcommand{\vect}[1]{\mathbf{#1}} 

\usepackage{comment}
\usepackage{subcaption}

\usepackage{pstricks}
\usepackage[countmax]{subfloat}
\usepackage{setspace}
\usepackage{hyperref}
\usepackage{color}
\usepackage{booktabs,multirow}
\usepackage{enumitem}
\usepackage{mathtools}
\usepackage{amssymb}
\usepackage{tikz}
\usepackage{booktabs,multirow}
\usetikzlibrary{patterns,arrows}
\usepackage{dsfont}

\usepackage{cite}

\makeatletter

\newcommand{\Rmnum}[1]{\expandafter\@slowromancap\romannumeral #1@}
\makeatother

\newtheorem{definition}{Definition}

\newcommand{\undertriangleleft}{\underline{\triangleleft}}

\providecommand{\propositionname}{Proposition}
\usepackage{eqparbox}

\usepackage[T1]{fontenc}
\usepackage{etoolbox}

\begin{document}

\title{SheafAlign: A Sheaf-theoretic Framework for Decentralized Multimodal Alignment}
\author{
	\IEEEauthorblockN{Abdulmomen Ghalkha, Zhuojun Tian, Chaouki Ben Issaid, and Mehdi Bennis\\
	}
    \thanks{
A.~Ghalkha, Z.~Tian, C.~B.~Issaid, and M.~Bennis are with the Center for Wireless Communications, University of Oulu, Oulu 90014, Finland. Email: \{abdulmomen.ghalkha, zhuojun.tian, chaouki.benissaid, mehdi.bennis\}@oulu.fi.
}
    }

\maketitle

\begin{abstract}
Conventional multimodal alignment methods assume mutual redundancy across all modalities, an assumption that fails in real-world distributed scenarios. We propose SheafAlign, a sheaf-theoretic framework for decentralized multimodal alignment that replaces single-space alignment with multiple comparison spaces. This approach models pairwise modality relations through sheaf structures and leverages decentralized contrastive learning-based objectives for training. SheafAlign overcomes the limitations of prior methods by not requiring mutual redundancy among all modalities, preserving both shared and unique information. Experiments on multimodal sensing datasets show superior zero-shot generalization, cross-modal alignment, and robustness to missing modalities, with 50\% lower communication cost than state-of-the-art baselines.
\end{abstract}


\input{introduction}

\input{system_model}

\input{proposed_algorithm}

\input{simulation_results}

\small
\bibliographystyle{IEEEtran}
\bibliography{references} 

\end{document}

%% file: introduction.tex
\section{Introduction}
The rapid proliferation of multimodal sensor networks and distributed data collection systems mandates the development of learning frameworks capable of effectively integrating heterogeneous sources of information. Multimodal learning aims to integrate information from heterogeneous sources, such as images, audio, LiDAR, and wireless signals, for understanding a common event of interest. Central to this goal is multimodal alignment, where the learned embeddings must be semantically consistent across different modalities, facilitating downstream tasks. 

Existing multimodal frameworks such as CLIP \cite{radford2021learning}, AudioCLIP \cite{guzhov2022audioclip}, and ImageBind \cite{girdhar2023imagebind} embed all modalities into a single, shared embedding space. In CLIP, alignment is established between images and text through contrastive learning. AudioCLIP extends this alignment to audio by adapting the CLIP framework to audio-text-image triplets. Similarly, ImageBind generalizes this further by binding six modalities (images, text, audio, depth, thermal, IMU) into one shared embedding space, relying solely on image-paired data to induce alignment.

While these unified embedding spaces enable strong performance in multimodal tasks, they have significant limitations, particularly in distributed settings where sensors have partially overlapping information. Collapsing all modalities into a single space can suppress modality-specific information and fail to capture nuanced relationships between certain modality pairs. Specifically, visual bias as binding all modalities via images risks distorting non-visual semantics, leading to suboptimal pairings since direct modality links (e.g., audio-text) are overlooked. Furthermore, in scenarios where certain modalities are absent, models trained within such a single space often struggle to infer or reconstruct missing information, particularly when the alignment relies on modalities with insufficient mutual redundancy or is biased towards dominant modalities such as vision. The key challenge is to move beyond these rigid global alignments and design a framework that can model cross-modal relationships, maintaining alignment and inference capabilities, when modalities exhibit limited redundancy or partial observability.

To address this challenge, we propose SheafAlign, a sheaf-theoretic multimodal alignment framework that introduces comparison spaces, latent spaces where embeddings from different modalities can be projected and directly aligned.  Unlike single-space alignment approaches, SheafAlign models pairwise modality interactions using sheaves defined over a communication graph, where each edge corresponds to a shared comparison space between the two nodes it connects. Building on the formulation in \cite{issaidtackling, ghalkha2025sheaf} and defining a sheaf over embeddings, this structure enables modality pairs to align locally without assuming mutual redundancy across all modalities, allowing richer and more pair-specific relationships. Furthermore, SheafAlign does not require the presence of a specific reference modality across all samples, enhancing robustness to missing or partially observed data during training and inference. The main contributions of this article are summarized as follows
\begin{itemize}
     \item We introduce SheafAlign, a novel sheaf-theoretic framework for multimodal alignment that replaces a single embedding space with a network of comparison spaces. 
    \item We develop a novel decentralized training procedure that enables clients to collaboratively learn local modality-specific embeddings and pairwise relations, enhancing scalability and communication efficiency.
    \item Experiments on real-world and synthetic multimodal datasets demonstrate superior cross-modal alignment, zero-shot and few-shot generalization, and robustness to missing modalities, while achieving up to 50\% communication savings compared to other baselines. 

\end{itemize}
The rest of this paper is organized as follows. Section \ref{System_Model} presents the system model and problem formulation. Section \ref{proposed_algorithm} introduces the proposed sheaf-theoretic framework for multimodal alignment. In Section \ref{Simulation_section}, we evaluate the performance of SheafAlign through real-world and synthetic datasets.

%% file: system_model.tex
\section{System Model and Problem Formulation}\label{System_Model}

In the decentralized learning framework, the system is represented as a connected graph $\mathcal{G} = (\mathcal{V}, \mathcal{E})$, where each node $i \in \mathcal{V} = \{1, \ldots, N\}$ represents a client equipped with a single modality $m_i \in \mathcal{M} = \{1, \ldots, M\}$. The edges, denoted by $\mathcal{E}$, define the direct communication links between clients. Each client $i$ collects its local observations $\bm{x}_{m_i}$ with its modality $m_i$. 

All clients are assumed to observe the same underlying event or scene. We do not assume access to labels for these observations. Instead, clients learn collaboratively by exchanging information with their neighbors through the communication graph $\mathcal{G}$, allowing the system to discover relationships between modalities and capture cross-modal dependencies without centralized supervision. Each client applies a modality-specific feature extractor $f_i(\cdot)$ to obtain its embedding $\bm{h}_i = f_i(\bm{x}_{m_i})$, to learn aligned embeddings across modalities that satisfy $\bm{h}_i \approx \bm{h}_j, \forall i, j \in \mathcal{V}$, if $\bm{x}_{m_i}$ and $\bm{x}_{m_j}$ originate from the same sample.

A key challenge in decentralized multimodal alignment is achieving robust aligned representations across distributed sensors with heterogeneous modalities. Existing frameworks address this by embedding all modalities into a single shared embedding space, implicitly assuming equal information redundancy across modalities. However, in distributed settings, this assumption rarely holds as spatially separated sensors often capture views with different levels of overlap and correlation. As a result, visual bias, loss of modality-specific information, and weak alignment between non-visual pairs can occur, limiting fine-grained semantic consistency across modalities. To better illustrate these limitations, following the definition in \cite{liang2023factorized}, we define the information-theoretic decomposition of task-relevant information across modalities.

\begin{definition}[Information-Theoretic Decomposition]
\label{def_information_decomposition}
Let $\{\bm{X}_1, \dots, \bm{X}_M\}$ be a set of random variables representing $M$ modalities observing a common phenomenon, and let $Y$ denote a downstream task variable. The total task-relevant information can be decomposed as
\begin{align}
\nonumber
I(\bm{X}_1, \dots, \bm{X}_M; Y) = & \sum_{i} U_i + \sum_{i<j} R_{ij} + \dots + R_{1\dots M} \\
& \quad + \sum_{i<j} S_{ij} + \dots + S_{1\dots M},
\end{align}
where $U_i$ denotes the information about $Y$ that is unique to modality $\bm{X}_i$, while $R_{ij\dots k}$ and $S_{ij\dots k}$ represent, respectively, the redundant and synergistic information about $Y$ that is shared among modalities $\{\bm{X}_i, \bm{X}_j, \dots, \bm{X}_k\}$ but not by any subset thereof.
\end{definition}

For each modality $m_i$, a modality-specific encoder $f_i: \bm{X}_{m_i} \to \bm{h}_i$ maps the raw observation $\bm{X}_{m_i}$ to a learned embedding $\bm{h}_i = f_i(\bm{X}_{m_i})$. 
The goal of representation learning is to learn $\{f_i\}_{i=1}^M$ such that, as in Definition~\ref{def_information_decomposition}, the mutual information $I(\bm{h}_1, \ldots, \bm{h}_M; Y)$ is maximized. 
However, when all embeddings are constrained to a single shared space, the retained information about $Y$ is bounded by the smallest intersection across modalities, i.e., 
$I(\bm{h}_1, \ldots, \bm{h}_M; Y) \le R_{1 \ldots M}$, as shown in the bottom part of Figure~\ref{fig:algorithm}.
As a result, pairwise-shared and unique modality information cannot be fully captured, thereby hindering downstream task performance.

%% file: proposed_algorithm.tex
\section{Proposed Algorithm}\label{proposed_algorithm}
\begin{figure}
    \centering
    \includegraphics[width=0.9\linewidth]{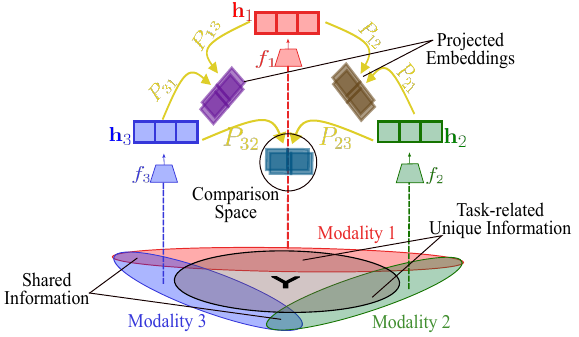}
    \caption{\textbf{Bottom:} Illustration of the absence of mutual redundancies between modalities. \textbf{Top:} Schematic diagram of SheafAlign architecture denoting restriction maps and comparison space alignment.}
    \label{fig:algorithm}
\end{figure}

To address the limitations of single-space alignment, we introduce multiple comparison spaces and adopt a pairwise contrastive learning strategy. With this, we allow each modality pair to align in its most informative subspace, rather than forcing all modalities into one shared embedding space. This approach captures shared information implicitly through overlapping pairwise comparisons while preserving non-redundant information. 
This formulation aligns naturally with sheaf theory, which provides a rigorous framework for representing local modality embeddings as locally consistent assignments and enforcing global consistency through the underlying sheaf structure. Inspired by that, we utilize the sheaf structure into the framework design, as illustrated in the top part of Figure~\ref{fig:algorithm}.

\subsection{Sheaf Structure}

A sheaf structure, as described in \cite{issaidtackling}, provides a principled framework for modeling interactions across decentralized agents by formalizing \textit{local views} and their pairwise relationships. 
Building on this idea, we shift the focus from modeling relations between model parameters to \textit{embeddings}. Specifically, we define a \textit{cellular sheaf over the embedding space}, enabling pairwise alignment of embeddings between modalities while maintaining unique information of each modality, thus mitigating limitations of forcing all embeddings into a single shared space.

Intuitively, a sheaf structure assigns a vector space to each node and edge in the network, along with linear maps that relate them. In our context, these vector spaces represent \textit{semantic embeddings}—intermediate feature representations extracted from local modality-specific observations. The sheaf structure captures how these local embeddings can be translated and compared between neighbors in the communication graph.

Formally, a \textit{cellular sheaf} $\mathcal{F}$ of $\mathbb{R}$-vector spaces over a communication graph $\mathcal{G} = (\mathcal{V}, \mathcal{E})$ consists of
\begin{itemize}
  \item For each node $i \in \mathcal{V}$, a local embedding space $\mathcal{F}(i) = \mathbb{R}^{d_i}$.
  \item For each edge $e = (i, j) \in \mathcal{E}$, a shared comparison space $\mathcal{F}(e) = \mathbb{R}^{d_{ij}}$.
  \item For each edge $e$ and incident vertex $i$, a linear \textit{restriction map} $\mathcal{F}_{i \undertriangleleft e}: \mathcal{F}(i) \to \mathcal{F}(e)$, projecting local embeddings onto a shared comparison space for alignment.
\end{itemize}

Here, the stalk $\mathcal{F}(i)$ represents the \textit{embedding space} of client $i$ (e.g., $\bm{h}_i \in \mathbb{R}^{d_i}$), and the maps $\mathcal{F}_{i \undertriangleleft e}$ are used to project embeddings into a shared comparison space for \textit{semantic comparison across neighbors}. Given an edge $(i, j)$, we denote the matrix representation of $\mathcal{F}_{i \undertriangleleft e}$, with
respect to a chosen basis such as the standard basis, by $\bm{P}_{ij}$, and let $\bm{P}_{ij}^T$ represent its transpose.  

Building on this sheaf-theoretic formulation, local alignment between neighboring embeddings is realized by projecting each embedding into the corresponding shared comparison space, $\mathcal{F}(e)$ via $\mathcal{F}_{i \undertriangleleft e}(\bm{h}_i)$ and $\mathcal{F}_{j \undertriangleleft e}(\bm{h}_j)$. The discrepancy between neighboring embeddings is captured by the \textit{sheaf Laplacian} $L_{\mathcal{F}}$, which measures the degree of consistency across all local assignments. The sheaf Laplacian is a block matrix, with each block $L_{ji}$ defined as
\begin{align}
    L_{ji} &= \begin{cases} 
    \sum_{i \undertriangleleft e} \mathcal{F}^*_{i \undertriangleleft e} \circ \mathcal{F}_{i \undertriangleleft e}, & \text{if } i = j, \\
    -\mathcal{F}^*_{j \undertriangleleft e} \circ \mathcal{F}_{i \undertriangleleft e}, & \text{if } e = (i, j) \in \mathcal{E}, \\
    \bm{0}, & \text{otherwise}.
    \end{cases}
\end{align}

The quadratic form $\bm{h}^\top_n L_{\mathcal{F}} \bm{h}_n$ quantifies how inconsistent the neighboring embeddings are across the network
\begin{align}
\label{eq_sheaf_lacplacian_loss}
\bm{h}^\top_n L_{\mathcal{F}} \bm{h}_n = \sum_{e = (i,j) \in \mathcal{E}} \mathds{1}_{(i, j), n} \| \mathcal{F}_{i \undertriangleleft e}(\bm{h}_{i, n}) - \mathcal{F}_{j \undertriangleleft e}(\bm{h}_{j, n}) \|^2.
\end{align}
where $\bm{h}_n = \{\bm{h}_{i,n}\}_{i \in \mathcal{V}}$, and $\mathds{1}_{(i,j),n}$ is 1 if both modalities $i$ and $j$ are observed for sample $n$, and 0 otherwise. This introduces an additional degree of freedom by allowing comparisons in dedicated edge-specific comparison spaces, while simultaneously preserving relationships and alignments between the original local embeddings.

\subsection{Sheaf-enabled Multimodal Alignment}

While the sheaf Laplacian enforces structural agreement, it alone cannot ensure that embeddings are semantically discriminative. To achieve this, we introduce a contrastive loss that operates directly within each shared comparison space $\mathcal{F}(e)$. Specifically, for an edge $e=(i,j)$ and a batch of $B$ co-observed samples, we first obtain the local embeddings $\{\bm{h}_{i,n}\}_{n=1}^B$ and $\{\bm{h}_{j,n}\}_{n=1}^B$.  These are then projected into the shared comparison space through $\bm{p}_{i,n}^{(e)} = \bm{P}_{i, j}\bm{h}_{i,n}$ and $\bm{p}_{j,n}^{(e)} = \bm{P}_{j, i}\bm{h}_{j,n}$.

We then apply an InfoNCE-style contrastive loss, treating $(\bm{p}_{i,n}^{(e)}, \bm{p}_{j,n}^{(e)})$ as a positive pair. The loss for edge $e$ is then defined as

\begin{align} \label{eq_contrastive_loss}
    \mathcal{L}_{\text{contrast}}^{(e)} = -\frac{1}{B} \sum_{n=1}^{B} \mathds{1}_{e, n}\log\frac{\exp(\text{sim}(\bm{p}_{i,n}^{(e)}, \bm{p}_{j,n}^{(e)})/\tau)}{\sum_{m=1}^{B}\exp(\text{sim}(\bm{p}_{i,n}^{(e)}, \bm{p}_{j,m}^{(e)})/\tau)},
\end{align}
where $\text{sim}(\cdot,\cdot)$ is the cosine similarity and $\tau$ is a temperature parameter. Following~\cite{oord2018cpc,poole2019variational}, minimizing the InfoNCE-style contrastive loss $\mathcal{L}_{\text{contrast}}^{(e)}$ over projected embeddings $\bm{p}_i^{(e)}$ and $\bm{p}_j^{(e)}$ is equivalent to maximizing a lower bound on their mutual information, $I(\bm{p}_i^{(e)}; \bm{p}_j^{(e)})$. Such alignment in comparison spaces over the sheaf is shown in the top of Figure~\ref{fig:algorithm}.

To improve cross-modal flexibility, we augment each edge $e=(i,j)$ with a dual map $\bm{Q}_{ij}: \mathcal{F}(e) \to \mathcal{F}(i)$, as a complement to the restriction map $\bm{P}_{ij}$, allowing us to avoid orthogonality assumptions on $\bm{P}_{ij}$. This dual map reconstructs local embeddings from shared comparison space projections, allowing inference of missing modalities $\tilde{\bm{h}}_{j \to i, n} = \bm{Q}_{ij}\big(\bm{P}_{ji}(\bm{h}_{j,n})\big)$.
The dual maps are trained via the reconstruction loss at node $i$ from neighbor $j$
\begin{align}
    \label{eq_reconstruction_loss}
   \mathcal{L}_{\text{recon}}^{(i,j)} = \frac{1}{B}\sum_{n=1}^B \mathds{1}_{(i, j), n} \big\| \bm{Q}_{ij}\bm{P}_{ji}(\bm{h}_{j,n}) - \bm{h}_{i,n} \big\|^2.
\end{align}

The final objective thus extends the sheaf Laplacian and contrastive alignment terms by incorporating bidirectional mappings between node and shared comparison spaces. In particular, the total loss is formally expressed as: 
\begin{align}
\label{eq_total_Loss}
    \mathcal{L}_{\text{total}} 
        = \lambda \sum_{n=1}^B \bm{h}_n^\top L_{\mathcal{F}} \bm{h}_n 
        + \beta \sum_{e \in \mathcal{E}} \mathcal{L}_{\text{contrast}}^{(e)}
        + \gamma \sum_{e\in\mathcal{E}} \mathcal{L}_{\text{recon}}^{(e)},
\end{align}
where $(\lambda, \beta, \gamma)$ are hyperparameters that balance consistency, discriminative alignment, and reconstruction of modality embeddings.

\subsection{Training Procedure}
The training process proceeds in a fully decentralized manner. Each node updates its parameters using only local data and the embeddings received from its immediate neighbors. 
 
In each iteration, every node $i$ computes its local embeddings $\bm{h}_{i,n} = f_i(\bm{x}_{{m_i},n})$, projects them onto the incident comparison spaces and exchanges these projections with neighboring nodes to evaluate the local loss $\mathcal{L}_{i,n}$. The node then computes the gradients of its loss $\mathcal{L}_i$, and updates its parameters $\bm{\theta}_i$, $\bm{P}_{ij}$, and $\bm{Q}_{ij}$ using gradient descent.

Following this procedure, all the parameters, including restriction maps $\bm{P}_{ij}$ and dual maps $\bm{Q}_{ij}$, are jointly learned without requiring any central coordinator. The entire end-to-end procedure, including embedding exchange, loss computation, and decentralized parameter updates, is summarized in Algorithm~\ref{alg:sheaf_learning}.

\begin{algorithm}[t]
\caption{SheafAlign: Decentralized Sheaf-Theoretic Multimodal Alignment}
\label{alg:sheaf_learning}
\begin{algorithmic}[1]
\State \textbf{Input:} Communication graph $\mathcal{G}=(\mathcal{V},\mathcal{E})$, epochs $T$, learning rate $\eta$, batch size $B$, hyperparameters $\lambda,\beta,\gamma$, temperature $\tau$.
\State \textbf{Initialize:} Feature extractors $\{\bm{\theta}_i\}_{i\in\mathcal{V}}$, restriction maps $\{\bm{P}_{ij}\}_{(i,j)\in\mathcal{E}}$, dual maps $\{\bm{Q}_{ij}\}_{(i,j)\in\mathcal{E}}$.

\For{epoch $=1$ to $T$}
    \State \textbf{Local step:} Each node $i\in\mathcal{V}$ computes embeddings $\vect{h}_{i}=f_i(\vect{x}_i;\bm{\theta}_i)$ from local data.
    \State \textbf{Edge alignment:} For each edge $(i,j)\in\mathcal{E}$ (in parallel):
        \State Project embeddings to comparison space $\mathcal{F}(e)$: 
        $\vect{p}_i^{(e)}=\bm{P}_{ij}\vect{h}_i$, $\vect{p}_j^{(e)}=\bm{P}_{ji}\vect{h}_j$.
        \State Nodes $i$ and $j$ exchange $\vect{p}_i^{(e)}$ and $\vect{p}_j^{(e)}$.
        \State Compute edge loss $\mathcal{L}^{(e)}=\lambda\mathcal{L}_{\text{Lap}}^{(e)}+\beta\mathcal{L}_{\text{contrast}}^{(e)}+\gamma\mathcal{L}_{\text{recon}}^{(e)}$ and the per node loss $\mathcal{L}_i$
        using (\ref{eq_sheaf_lacplacian_loss}, \ref{eq_contrastive_loss}), and (\ref{eq_reconstruction_loss}).
        \State \textbf{Parameter update:} 
        Each node $i$ updates $\bm{\theta}_i$, $\bm{P}_{ij}$, and $\bm{Q}_{ij}$ with gradient descent.
\EndFor
\end{algorithmic}
\end{algorithm}

This formulation preserves pairwise shared ($R_{ij}$), synergetic ($S_{ij}$), and unique ($U_i$) information, in contrast to conventional alignment methods, as discussed in section \ref{System_Model}. This is due to the fact that restriction maps $\bm{P}_{ij}$ act as filters extracting relevant features of each modality's embedding for alignment in the shared comparison space.

%% file: simulation_results.tex
\section{Numerical Results}\label{Simulation_section} 

\subsection{Training Settings}

We evaluate the proposed SheafAlign framework on three multimodal datasets: 
\begin{itemize}
\item \textbf{DeepSense Multimodal Blockage Prediction}~\cite{wu2023proactively}.  
A mmWave communication dataset capturing transmitter–receiver links under vehicular blockages. It includes three sensing modalities—RGB images, 2D LiDAR scans, and RF power measurements—for classifying whether the link is blocked.  For the feature encoders $f_i$, we use the same architectures described in \cite{ghalkha2025sheaf}.

\item \textbf{Multi-view MNIST}.  
A synthetic dataset derived from MNIST digits with three transformed views (original, edge-filtered, pixel-inverted), used to study multimodal alignment under controlled redundancy levels. We use a feature encoder consisting of two convolutional layers followed by a fully connected layer.

\item \textbf{Semantic Inpainting}~\cite{nishio2025semantic}.  
It consists of images from three cameras and CSI data from nine sensors, each providing different views of the same environment with varying redundancy across modalities. A pre-trained ResNet50 architecture is used.

\end{itemize}
For the last two datasets, we use images with different views to mimic modalities with varying levels of redundancy. For each dataset $N=3$ nodes are considered, and the communication graph is fully connected.
To evaluate SheafAlign, we compare it with two baselines: (i) ImageBind \cite{girdhar2023imagebind}, a state-of-the-art method aligning different modalities in a shared image embedding space; and (ii) a fully supervised learning baseline leveraging ground-truth labels during training. Model performance is assessed using zero-shot and few-shot accuracy (where the number of shots refers to the number of labeled samples), cross-modal retrieval recall (as defined in \cite{guzhov2022audioclip, girdhar2023imagebind}), and inference communication cost. 

For all experiments, we use a batch size of $B = 128$ and train for 50 epochs. The temperature constant in the contrastive loss is set to $\tau = 0.1$, which is a standard choice. Adam optimizer is used with a learning rate of $\eta = 10^{-3}$. For SheafAlign, the training parameters are set as $\lambda = 1.0$, $\beta = 1.0$, and $\gamma = 0.1$.

\subsection{Performance Comparison} 

\begin{figure*}[t]
    \centering
    \begin{subfigure}[t]{0.46\linewidth}
        \centering
        \includegraphics[width=\linewidth]{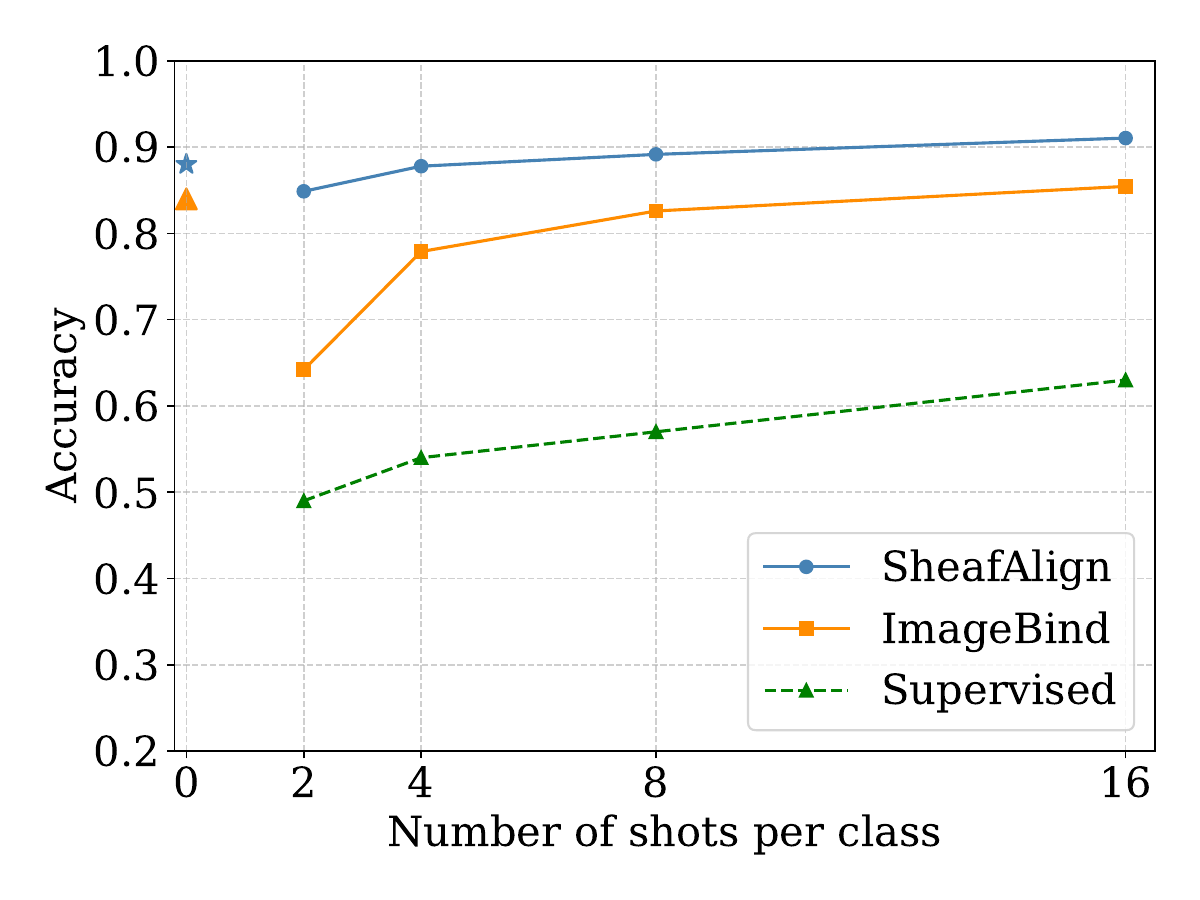}
        \caption{DeepSense blockage prediction.}
        \label{fig:zeroshot_blockage}
    \end{subfigure}
    \hfill
    \begin{subfigure}[t]{0.46\linewidth}
        \centering
        \includegraphics[width=\linewidth]{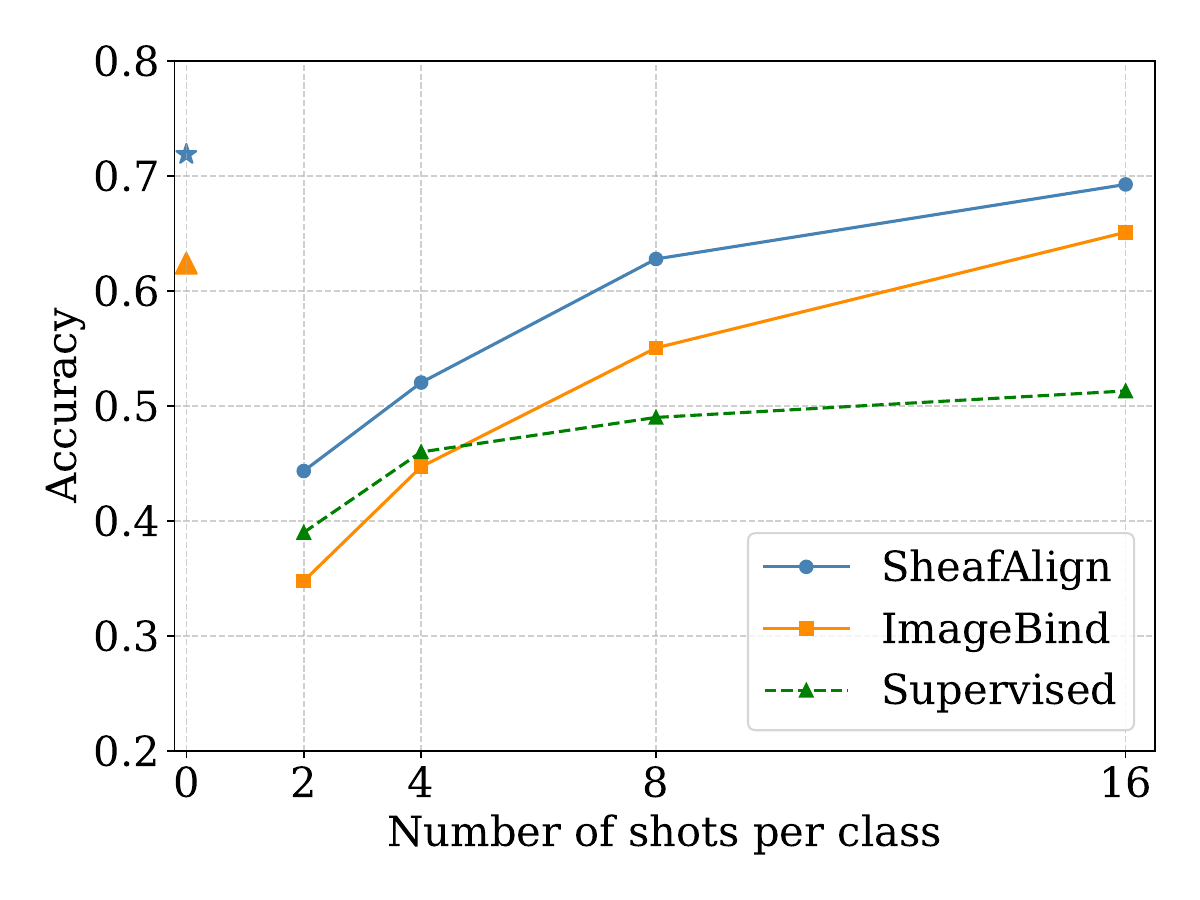}
        \caption{Multi-view MNIST.}
        \label{fig:zeroshot_mnist}
    \end{subfigure}

    \caption{Zero-shot and few-shot test accuracy for SheafAlign compared to ImageBind and supervised baselines across (a) DeepSense blockage prediction, and (b) Multi-View MNIST datasets, with zero-shot accuracies denoted as $\star$ and $\blacktriangle$ for SheafAlign and ImageBind, respectively.}
    \label{fig:zeroshot_combined}
\end{figure*}
Figure~\ref{fig:zeroshot_combined} demonstrates the superior performance of SheafAlign compared to ImageBind and the supervised baselines in zero- and few-shot settings. 
For the DeepSense blockage prediction dataset (Figure~\ref{fig:zeroshot_blockage}), accuracy saturates above four shots or samples per class, which can be attributed to the binary nature and simplicity of the task. In contrast, for the Multi-view MNIST dataset (Figure~\ref{fig:zeroshot_mnist}), the accuracy continues to improve with the number of labeled examples, indicating that SheafAlign effectively leverages limited supervision for more complex tasks. 
In both cases, SheafAlign consistently outperforms the baseline methods, maintaining 5\% accuracy improvement over the strongest baseline. The supervised models exhibit weaker generalization under limited data, reflecting their dependence on large labeled datasets. These results highlight that the embeddings produced by SheafAlign are both semantically meaningful and readily transferable to downstream tasks, even when only a small amount of labeled data is available for fine-tuning.

\begin{figure*}[t]
    \centering
    \begin{subfigure}[t]{0.46\textwidth}
        \centering
        \includegraphics[width=\linewidth]{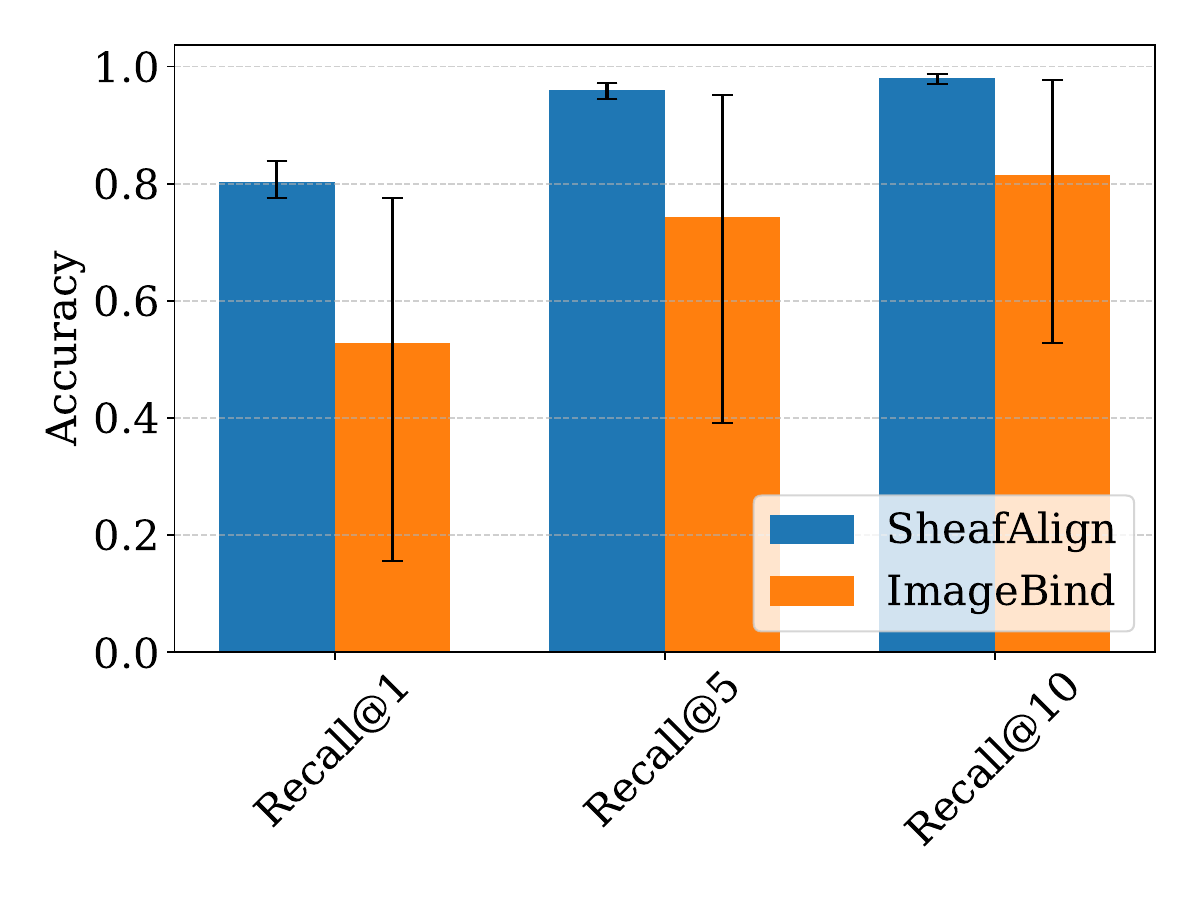}
        \caption{Multi-view MNIST dataset.}
        \label{fig:cross_mnist}
    \end{subfigure}
    \hfill
    \begin{subfigure}[t]{0.46\textwidth}
        \centering
        \includegraphics[width=\linewidth]{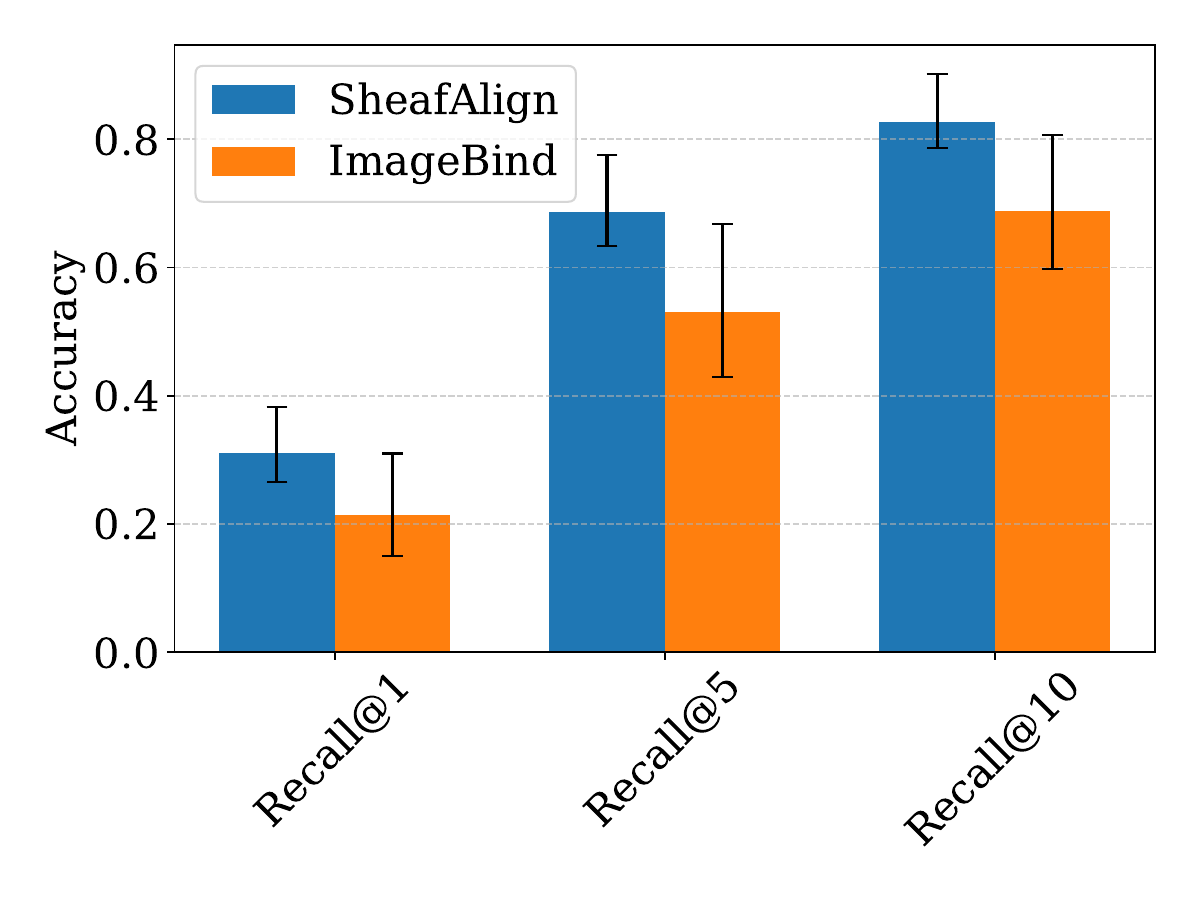}
        \caption{Semantic Inpainting dataset.}
        \label{fig:cross_semcom}
    \end{subfigure}

    \caption{Cross-modal retrieval performance of SheafAlign compared to ImageBind across (a) Multi-view MNIST, and (b) semantic inpainting datasets. 
    Bars show the mean Recall@K ($K = 1, 5, 10$), averaged across all modality pairs.}
    \label{fig:cross_combined}
\end{figure*}
The experiment results presented in Figure~\ref{fig:cross_combined} evaluate how well the learned embeddings are semantically aligned across modalities. Both datasets consist of multiple views and provide scenarios with varying degrees of redundancy. As shown in Figure~\ref{fig:cross_mnist}, SheafAlign achieves an average Recall@1 score that is 20\% higher than ImageBind, and an 18\% improvement in Recall@10. Similarly, Figure~\ref{fig:cross_semcom} demonstrates that SheafAlign attains more than 10\% higher recall scores. These results indicate that approaches relying on a single embedding space may yield high recall for certain modality pairs at the expense of others, limiting their ability to consistently capture well-aligned semantic correspondences, particularly when mutual redundancy is insufficient or uneven across modalities. Overall, these findings highlight the superiority of SheafAlign over conventional multimodal alignment frameworks.

\begin{table}[b]
    \centering
    \caption{Comparison of Average Accuracy and Communication Cost with sensors with dropout probability.} \label{tab:inference_comparison}
    \begin{tabular}{|c|c|c|c|}
        \hline
        $P_{\text{drop}}$ & Algorithm & Accuracy &  Transmitted Bytes [KB] \\ \hline
        0.1 & SheafAlign      & 0.87 & 46.20 \\ \hline
        0.1 & ImageBind  & 0.83 & 92.48 \\ \hline
        0.01 & SheafAlign     & 0.90 & 4.86  \\ \hline
        0.01 & ImageBind  & 0.84 & 9.72  \\ \hline
\end{tabular}
\end{table}

Finally, we evaluate the communication cost incurred when inferring missing embeddings under scenarios where sensors serving downstream tasks may drop with dropout probabilities $P_{\text{drop}} \in \{0.01, 0.1\}$. Semantically aligned embeddings can act as backup representations, allowing available sensors to share their embeddings to substitute the missing modality and maintain uninterrupted performance, particularly in real-time applications. As shown in Table~\ref{tab:inference_comparison}, SheafAlign achieves comparable accuracy while reducing communication overhead by approximately 50\%. This efficiency arises naturally from the sheaf architecture, as the comparison space is designed to be half the size of the original embedding space. Consequently, the sheaf structure emphasizes extracting the most relevant information for each modality pair, while retaining other unique details that may be useful for alternative pairs. These properties make SheafAlign particularly promising for deployment in resource-constrained or latency-sensitive applications.